\documentclass{article}
\pdfoutput=1
\usepackage[nonatbib, final]{nips_2016}
\usepackage{tikz}
\usepackage[subpreambles=true]{standalone}
\usepackage{import}
\usepackage{footnote}
\usepackage[square,numbers]{natbib}
\newcommand{\eref}[1]{(\ref{#1})}
\newcommand{\aref}[1]{Algorithm~\ref{#1}}
\newcommand{\sref}[1]{Section~\ref{#1}}
\newcommand{\figref}[1]{Figure~\ref{#1}}

\usepackage[utf8]{inputenc} 
\usepackage[T1]{fontenc}    
\usepackage{hyperref}       
\usepackage{url}            
\usepackage{booktabs}       
\usepackage{amsfonts}       
\usepackage{nicefrac}       
\usepackage{microtype}      

\usepackage[utf8]{inputenc}
\usepackage{indentfirst}
\usepackage{amsmath}
\usepackage{graphicx}
\usepackage{amssymb}
\usepackage{bm} 
\usepackage{mathtools}

\usepackage{algorithm}
\usepackage{caption}
\usepackage{subcaption}
\usepackage{graphicx}
\usepackage{algorithmic}
\usepackage{booktabs}

\newcommand{\vi}[1]{\mathbf{#1}}
\newcommand{\M}[1]{\mathsf{#1}}

\newcommand{\state}{\vi{x}}
\newcommand{\dstate}{\delta\vi{x}}

\newcommand{\control}{\vi{u}}
\newcommand{\dcontrol}{\delta\vi{u}}

\newcommand{\inputstate}{\widetilde{\vi{x}}}

\newcommand{\brownian}{\vi{\bm{\omega}}}
\newcommand{\uncertainty}{\vi{\bm{\xi}}_\M{G}}
\newcommand{\stochasticity}{\vi{\bm{\xi}}_\M{F}}

\newcommand{\dt}{\delta t}

\DeclareMathOperator{\statedyn}{\vi{f}}
\DeclareMathOperator{\knowndyn}{\vi{f}_{sim}}
\DeclareMathOperator{\unknowndyn}{\vi{\epsilon}}
\DeclareMathOperator{\learnedunknowndyn}{\hat{\vi{\epsilon}}}

\DeclareMathOperator{\policy}{\pi}

\DeclareMathOperator{\noisedyn}{\M{F}}
\DeclareMathOperator{\uncertaintyCov}{\M{\Sigma}}
\DeclareMathOperator{\uncertaintydyn}{\M{G}}

\newcommand{\Dstate}{\Delta\state}

\newcommand{\A}{\vi{A}}
\newcommand{\B}{\vi{B}}

\newcommand{\C}{\M{C}}
\newcommand{\D}{\M{D}}

\newcommand{\cost}{{l}}

\newcommand{\lx}{\vi{l}_\state}
\newcommand{\lu}{\vi{l}_\control}
\newcommand{\lxx}{\vi{l}_{\state\state}}
\newcommand{\lux}{\vi{l}_{\control\state}}
\newcommand{\luu}{\vi{l}_{\control\control}}

\newcommand{\V}{{V}}

\newcommand{\tr}{tr}

\newcommand{\Ss}{\vi{S}}
\newcommand{\s}{\vi{s}}
\newcommand{\G}{\vi{G}}
\newcommand{\g}{\vi{g}}
\newcommand{\q}{\vi{q}}

\title{ GP-ILQG: Data-driven Robust Optimal Control for Uncertain Nonlinear Dynamical Systems}

\author{
  Gilwoo Lee 
   \And 
    Siddhartha S. Srinivasa\\ 
   \And 
    Matthew T. Mason \thanks{\texttt{gilwool, siddh, mason@cs.cmu.edu}. All authors are affiliated with Robotics Institute, Carnegie Mellon University.}
}

\begin{document}

\maketitle

\begin{abstract}
    As we aim to control complex systems, use of a simulator in model-based reinforcement learning is becoming more common. However, it has been challenging to overcome the Reality Gap, which comes from nonlinear model bias and susceptibility to disturbance. To address these problems, we propose a novel algorithm that combines data-driven system identification approach (Gaussian Process) with a Differential-Dynamic-Programming-based robust optimal control method (Iterative Linear Quadratic Control). Our algorithm uses the simulator's model as the mean function for a Gaussian Process and learns only the difference between the simulator's prediction and actual observations, making it a natural hybrid of simulation and real-world observation. We show that our approach quickly corrects incorrect models, comes up with robust optimal controllers, and transfers its acquired model knowledge to new tasks efficiently. 
\end{abstract}

\section{Introduction}
As we aim to control more complex robotic systems autonomously, simulators are being more frequently used for training in model-based reinforcement learning~\cite{kober2013reinforcement}. A simulator allows us to explore various policies without damaging the robot, and is also capable of generating a large amount of synthetic data with little cost and time. 

However, we often observe that simulator-based policies perform poorly in real world, due to model discrepancy between the simulation and the real world. This discrepancy arises from two fundamental challenges: (1) system identification to match the simulation model with the real world requires the exploration of a large state space at the risk of damaging the robot, and (2) even with good system identification, there is still discrepancy due to the limitations of a simulator's ability to render real-world physics.

Stochastic optimal control algorithms attempt to partially address this issue by artificially injecting noise into the simulation during training~\cite{huh2009real,wang2010optimizing}, or by   explicitly modeling multiplicative noise~\cite{todorov2005generalized, stoorvogel1992h_}.

If the task domain is predefined, exploration can be limited to task-specific trajectories, and system identification can be coupled with trajectory optimization~\cite{Tan-IROS-16}. Some recent works have suggested policy training with multiple models, which results in a policy robust to model variance~\cite{mordatch2016combining, boeing2012leveraging}. While these methods have shown some successful results, they are still limited by the expressiveness of the simulator's model. If the true model is outside of the simulator's model space, little can be guaranteed. 

Thus, although these algorithms produce more robust policies, they fail to address the fundamental issue: there is unknown but structured model discrepancy between the simulation and the real world.

\begin{figure}[t!]
\centering
\includegraphics{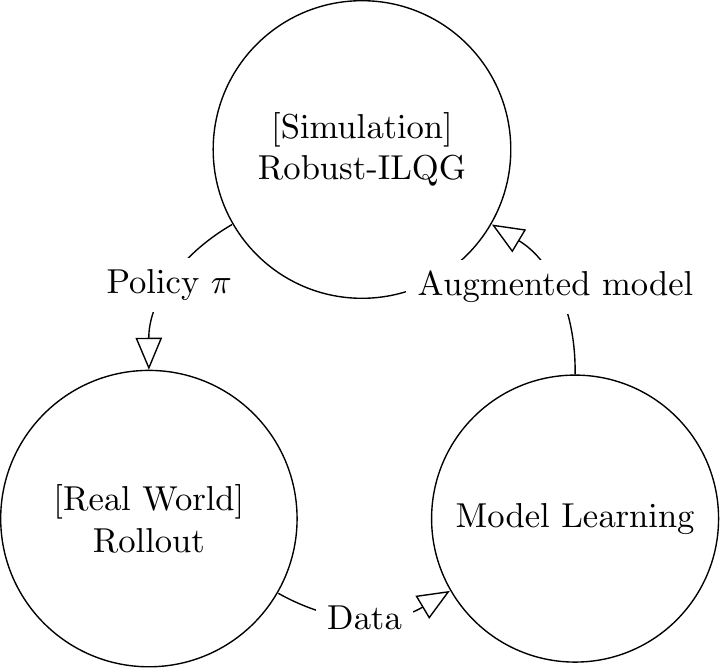}
\caption{GP-ILQG overview}
\label{fig:algorithm}
\end{figure}

In this paper, we propose a novel algorithm that addresses both model bias and multiplicative noise. Our work is based on the following key insight:
\begin{quote}
Explicitly correcting model bias and incorporating the correction as well as our uncertainty of the correction in optimal control enables lifelong learning of the system \emph{and} robust control under uncertainty.
\end{quote}

Our algorithm iterates over simulation-based optimal control, real-world data collection, and model learning, as illustrated in \figref{fig:algorithm}. Starting from a potentially incorrect model given by the simulator, we obtain a control policy, with which we collect data in the real world. This data feeds into model learning, during which we correct model bias and estimate our uncertainty of the correction. Both the correction and its uncertainty are incorporated into computing a robust optimal control policy, which then gets used to collect more data.   

Our approach improves any simulator beyond the scope of its model space to match real-world observations and produces an optimal control policy robust to model uncertainty and multiplicative noise. The improved simulator uses previous real-world observations to infer the true model when it explores previously visited space, but when it encounters a new region, it relies on the simulator's original model. Due to this hybrid nature, our algorithm shows faster convergence to the optimal policy than a pure data-driven approach~\cite{pan2014probabilistic} or a pure simulation-based approach. Moreover, as it permanently improves the simulator, it shows even faster convergence in new tasks in similar task domain. 

\section{Related Work}

Most model-based reinforcement learning has both model learning (system identification) and policy optimization components~\cite{kober2013reinforcement}. The data for a model comes either from real world or simulation, and is combined to construct a model via nonlinear function approximators such as Locally Weighted Regression~\cite{atkesonlocally}, Gaussian Processes~\cite{rasmussen2006gaussian}, or Neural Networks~\cite{narendra1990identification}. Once the model is built, a typical policy gradient method computes the derivatives of the cost function with respect to control parameters~\cite{deisenroth2011pilco, levine2013guided}. 

If an analytic model is given, e.g., via equations of motion or as a simulator\footnote{We consider black-box simulators as analytical models, as derivatives can be taken by finite differencing.}, one can use classical optimal control techniques such as Differential Dynamic Programming (DDP)~\cite{jacobson1970differential}, which compute a reference trajectory as well as linear feedback control law. For robustness, Iterative Linear Quadratic Gaussian Control (ILQG)~\cite{todorov2005generalized} or H-$\infty$ Control \cite{stoorvogel1992h_} can be used to incorporate multiplicative noise. Variants of \cite{todorov2005generalized} have been used to generate guiding policies for data-driven RL methods~\cite{levine2013guided, zhang2016learning}. Recently, there have been some attempts to combine DDP or ILQG with data-driven models by replacing analytical models with locally linear models~\cite{mitrovic2010adaptive, yamaguchi2015differential} or nonlinear models~\cite{pan2014probabilistic, PanNIPSWS15, yamaguchi2016neural, pan2015data} learned by Gaussian Processes or Neural Networks. 

The goal of our algorithm is closely aligned with those of \cite{zagal2004back} and \cite{icml2006_AbbeelQN06}. \cite{zagal2004back} has proposed a framework in which the robot maintains two controllers, one for the simulator and another for the real world, and aims to narrow the difference. \cite{icml2006_AbbeelQN06} assumes a deterministic real world, constructs an optimal policy based on the simulator's deterministic model, and evaluates its performance in the real world, while successively augmenting the simulator's model with time-dependent corrections based on real-world observations. Our method considers a stochastic system and the correction is not time-dependent.

\section{Approach}\label{sec:approach}

We work with stochastic nonlinear dynamical systems, whose evolution is described by the stochastic differential equation
\begin{equation}\label{eq:dx}
    d\state = \statedyn(\state, \control)\,dt
            + \noisedyn(\state, \control)\,d\brownian
\end{equation}
where the applied control $\control \in \mathbb{R}^m$ transitions 
the state $\state \in \mathbb{R}^n$ of the system according to a linear sum of the system dynamics $\statedyn \in \mathbb{R}^n$ and state-dependent amplification $\noisedyn \in \mathbb{R}^{n\times p}$ of Brownian noise increment $d\brownian \in \mathbb{R}^p$.

We assume that we have full knowledge of $\noisedyn$ and partial knowledge of $\statedyn$, e.g., from a simulator. We represent $\statedyn$ as the sum of simulated component $\knowndyn$ and unknown residual $\unknowndyn$,
\begin{equation*}
  \statedyn(\state, \control) = \knowndyn(\state, \control) + \unknowndyn(\state, \control).
\end{equation*}

Then, in a discrete setting with fixed step size $\dt$, \eref{eq:dx} can be rewritten as the following:
\begin{align}\label{eq:discrete-dx-uncertain}
    \Dstate = (\knowndyn + \epsilon)\dt + \noisedyn \stochasticity\sqrt{\dt} 
\end{align}
where $\stochasticity\sim \mathcal{N}(0, I_{p\times p})$ and $\sqrt{\delta t}$ appears because the noise covariance increases linearly with time. We omit $\state$ and $\control$ for clarity.

Nonzero model bias results in nonzero difference between the model's prediction $\knowndyn\dt$ and the actual $\dstate$. From \eqref{eq:discrete-dx-uncertain}, the difference is equivalent to 
\begin{align*}
    \Dstate - \knowndyn\dt = \unknowndyn\dt + \noisedyn \stochasticity\sqrt{\dt} 
\end{align*}
In expectation, 
\begin{align*}
    \mathbb{E}[\unknowndyn\dt] = \mathbb{E}[   \Dstate - \knowndyn\dt- \noisedyn \stochasticity\sqrt{\dt} ] =  \mathbb{E}[   \Dstate - \knowndyn\dt]
\end{align*}
as the mean of $\stochasticity$ is 0. 

In order to correct the nonzero model bias, we approximate $\unknowndyn\dt$ from data. From rollouts of trajectories, we collect a set of $\{\state, \control,  \Dstate - \knowndyn\dt\}$ tuples. Assuming that the the covariance of $\unknowndyn\dt$ grows linearly with $\dt$, we use a Gaussian Process (GP)~\cite{rasmussen2006gaussian}, which estimates $\unknowndyn\dt$ as a Gaussian distribution:
\begin{equation*}
    \unknowndyn(\state, \control)\dt \sim \mathcal{N}\big(\hat{\unknowndyn} (\state, \control)\dt , \uncertaintyCov(\state, \control)\dt\big)
\end{equation*}
whose derivation we provide in \sref{subsec:GP}. Let $\uncertaintyCov(\state, \control)$ be decomposed into $ \uncertaintydyn(\state, \control) \uncertaintydyn(\state, \control) ^\top $, with $\uncertaintydyn\in\mathbb{R}^{n\times n}$. Now \eqref{eq:discrete-dx-uncertain} can  be rewritten as the following: 
\begin{equation}\label{eq:dx-uncertainty}
\Dstate = (\knowndyn + \learnedunknowndyn)\,\dt + \noisedyn\stochasticity\sqrt{\dt} + \uncertaintydyn\,\uncertainty\sqrt{\dt}
\end{equation}
where $\uncertainty \sim \mathcal{N}(0,I_{n\times n})$  . 

The original ILQG handles stochastic systems without uncertainty, but the above formulation makes it straightforward to extend ILQG to incorporate uncertainty, which we refer to as Robust-ILQG~(see \sref{sec:ILQG}).

Our approach is summarized in \aref{alg1}. Initially, with zero data and no policy, we start by using ILQG\footnote{Robust-ILQG is equivalent to ILQG in the absence of $\unknowndyn$.} to obtain a locally optimal policy for the suboptimal model provided by $\knowndyn$. Then, with rollouts of the computed policy, we compute the error between $\statedyn$ and $\knowndyn$, with which we train a GP that estimates $\unknowndyn$. We run Robust-ILQG again, but with the updated dynamics. This process is repeated until the policy converges.

One major advantage of our approach is that it is straightforward to use the improved model for a new task. Given observations learned from a previous task, \aref{alg1} starts with an estimate of $\unknowndyn$. Whenever the algorithm explores a previously visited state-control space, the learner corrects any simulator error and provides smaller uncertainty covariance; in a new region, the algorithm relies on the original model $\knowndyn$ with larger uncertainty covariance than the explored region. This results in a more accurate model, and as Robust-ILQG takes into account uncertainty, it also results in a more robust policy than those that rely only on the simulator or on a locally learned dynamics. When using only a partially-correct simulated model, the policy generated is always limited by the simulator's accuracy; when using a locally learned dynamics, the robot knows very little outside previously explored regions. Our algorithm, which combines both, our algorithm quickly outperforms the simulator-only approaches and requires less data than the latter.

\begin{algorithm}[t!]
\caption{GP-ILQG}\label{alg1}
\begin{algorithmic}
\REQUIRE $\pi$, $\knowndyn$, $\noisedyn$, $\vi{D}$
\IF {$\vi{D} \neq \emptyset$}
\STATE $[\learnedunknowndyn, \uncertaintydyn] \leftarrow \texttt{GP}(\vi{D})$
\ENDIF
\IF {$\pi == NIL$}
\STATE $\pi \leftarrow \texttt{Robust-ILQG}(\knowndyn +\learnedunknowndyn, \noisedyn, \uncertaintydyn)$
\ENDIF

\WHILE {$\Delta \pi > \gamma$}
\STATE Collect $\{(\state, \control, \dstate - \statedyn\dt)\}$ with rollouts of $\policy$. 
\STATE $\vi{D} \leftarrow \vi{D} \cup  \{(\state, \control, \dstate - \statedyn\dt)\}$
\STATE $[\learnedunknowndyn, \uncertaintydyn] \leftarrow \texttt{GP}(\vi{D})$
\STATE $\pi \leftarrow \texttt{Robust-ILQG}(\knowndyn +\learnedunknowndyn, \noisedyn, \uncertaintydyn)$.
\ENDWHILE
\end{algorithmic}
\end{algorithm}

\begin{figure}[t!]
\centering
\includegraphics{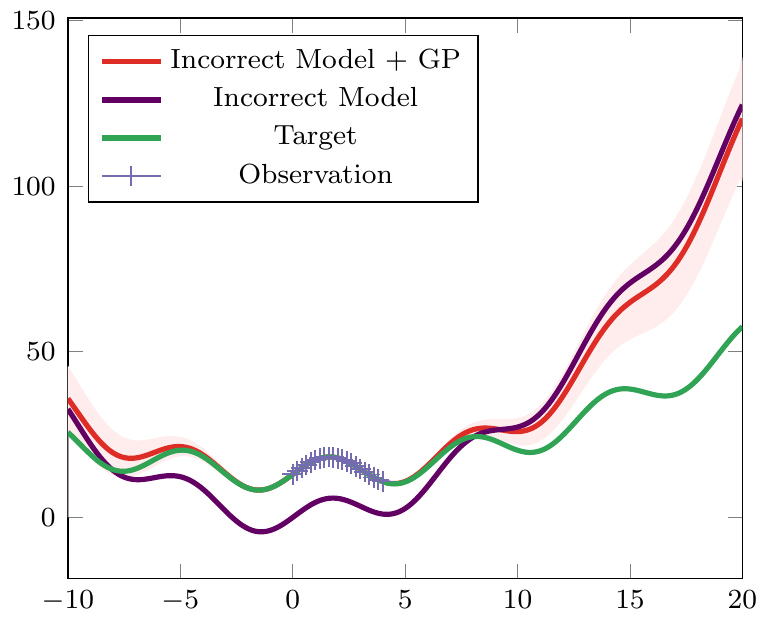}
\caption{Use of Gaussian Process to correct an incorrect model. Near previous observations, the model's prediction is corrected to match the target. When the test input is far from the prior observations, predictions resort to the incorrect model.  Shaded area indicates 95\% confidence interval.}
\label{fig:correctiveGP}
\end{figure}

\subsection{Gaussian Process with the Simulator as a Mean Function}\label{subsec:GP}

Gaussian Process Regression~\cite{rasmussen2006gaussian} is a nonlinear regression technique that has been successfully used in many model-based reinforcement learning approaches~\cite{deisenroth2011pilco,pan2014probabilistic}. A Gaussian Process is a stochastic process in which any finite subset is jointly Gaussian. Given a set of inputs and corresponding outputs $\{x_i, y_i\}_{i=1}^n$ the distribution of $\{y_i\}_{i=1}^n$ is defined by the covariance given by a kernel function $k(x_i, x_j)$. 

We use a variant of Gaussian Processes that uses a nonzero mean function. With $f:X \rightarrow Y$ as the mean function, the prediction for test input $x$ becomes the following:
\begin{align*}
    \mathbb{E}[y] &= f(x) + k_{xX}^\top K ^{-1}(Y-f(X)) \\
    var(y) &=k_{xx} - k_{xX}^\top K^{-1}k_{xX}
\end{align*}
where $k_{xX}$ is the covariance between test input $x$ and training set $X$, $K$ is the covariance matrix of among the elements of $X$. In this formulation, the GP provides a posterior distribution given $f(x)$ and observations.

Using the simulator as the mean function for a Gaussian Process allows us to combine the simulator and real-world observations smoothly. Near previous observations, the simulator's prediction is corrected to match the target. When the test input is far from the previous observations, the predictions resort to the simulator. See \figref{fig:correctiveGP} for an illustration.

As we have both $\state$ and $\control$ as the input, we define $\inputstate = [\state^\top \control^\top]^\top$ to be the input and $\delta \state$ to be the output, and use $\knowndyn$ as the mean function. Then, the GP serves to correct the error, $\delta \state - \knowndyn\delta t$. We use the ARD (Automatic Relevance Determination) squared exponential  kernel function:
\begin{align*}
    k(\inputstate_i,\inputstate_j) = \sigma_f^2 \exp(-(\inputstate_i - \inputstate_j)^\top \Lambda^{-1}(\inputstate_i - \inputstate_j))
\end{align*}
where $\sigma_f^2$ is the signal variance and $\Lambda$ controls the characteristic length of each input dimension. These hyperparameters are optimized to maximize log-likelihood of data using conventional optimization methods. For multiple dimensions of $Y$, we train one GP per dimension and treat each dimension to be independent.

\subsection{Robust-ILQG}\label{sec:ILQG}

In order to address both uncertainty and stochasticity in computing an optimal policy, we propose a new algorithm, which we refer to as Robust-Iterative Linear Quadratic Gaussian Control (Robust-ILQG). Robust-ILQG is an extension to the original ILQG~\cite{todorov2005generalized}, a variant of Differential Dynamic Programming (DDP) for stochastic systems. It makes second-order approximation of the value function to the second order and first-order approximation of the dynamics. The main difference between ILQG and Robust-ILQG is that the latter takes into account uncertainty in dynamics, as we will see in the following derivation.

We use a discretized version of a continuous system, with a fixed step size $\dt$. From the equation
\begin{align}
\begin{split}
    \Dstate &= \statedyn(\state, \control)\dt + \noisedyn(\state, \control)\sqrt{\dt}\stochasticity \\
    &= (\knowndyn + \learnedunknowndyn)\dt + \noisedyn\stochasticity\sqrt{\dt} + \uncertaintydyn\uncertainty\sqrt{\dt}
    \end{split}
\end{align}
The state $\state'$ at the next timestep can be written as 
\begin{align}\label{eq:discrete-f}
     \state'=  \state + (\knowndyn + \learnedunknowndyn)\dt + \noisedyn\stochasticity \sqrt{\dt}+ \uncertaintydyn\uncertainty\sqrt{\dt}
\end{align}

Given a trajectory $\{\state_0, \control_0, \cdots \state_T\}$, the total cost is given as 
\begin{align}\label{eq:total-cost}
    J(\state_0) = \mathbb{E}\bigg[\cost_T(\state_T) + \sum_{i=0}^{T-1} \cost(\state_i, \control_i) \bigg]
\end{align}
where $\cost_T$ is the final cost and $\cost$ is the running cost. Our objective is to find a deterministic policy $\pi:\vi{X} \rightarrow \vi{U} $ that minimizes this total cost.

We consider the local variation of value function with respect to change in $\state$ and $\control$. The Bellman equation gives us the following:
\begin{align}
\begin{split}\label{eq:bellman}
    \V(\state + \dstate) = \cost(\state + \dstate, \control + \dcontrol^*) + \mathbb{E}\bigg[\V'(\state' + \dstate'^*) \bigg]
\end{split}
\end{align}
where $\V$ is the value function at a particular timestep and $\V'$ is the value function at the next timestep. $\dstate'^*$ refers to the variation of the next state given an optimal control change $\dcontrol^*$.

Analogous to ILQR which does not have stochastic components, we can analytically derive the local optimal improvement $\dstate^*$ by taking first order approximation of dynamics and second order approximation of the value function.

The local deviation of the next state from its nominal (deterministic) state can be written as the following: 
\begin{align}
    \dstate' &= \A \dstate+ \B\dcontrol + \C \stochasticity +\D \uncertainty
\end{align}
where $\A, \B$ linearizes the deterministic  component of  dynamics at the current time step, defined as
\begin{align*}
 \A &= \mathbf{I} + (\knowndyn_\state + \learnedunknowndyn_\state)\dt\\
\B &= (\knowndyn_\control + \learnedunknowndyn_\control)\dt
\end{align*}
and $\C, \D$ captures the deviation from nominal (deterministic) state caused by stochastic terms 
\begin{align*}
    \C &= (\noisedyn + \noisedyn_\state\otimes
 \dstate + \noisedyn_\control\otimes \dcontrol)\sqrt{\dt}\\
    \D &= (\uncertaintydyn + \uncertaintydyn_\state\otimes \dstate + \uncertaintydyn_\control\otimes \dcontrol)\sqrt{\dt}.
\end{align*}
where $\otimes$ denotes tensor-vector multiplication as defined in \sref{sec:Appendix} such that $\noisedyn_\state\otimes \dstate$,  $\uncertaintydyn_\state\otimes \dstate$ are matrices of size $n\times p$ and $\noisedyn_\control\otimes \dcontrol$,  $\uncertaintydyn_\control\otimes \dcontrol$ are of size $n \times m$.

Assuming that the variation of value at the next timestep can be written in a quadratic form, 
\begin{align}
     V'(\state' + \dstate')  = s' + \s'^\top \dstate' +  \frac{1}{2}\dstate'^\top \Ss' \dstate'
\end{align}

the local variation of cost-to-go at the current time step can be written as \footnote{We assume that $\lux$ is zero.}
\begin{align*}
V(\state + \dstate) &= l(\state, \control)+ \lx^\top \dstate + \frac{1}{2}\dstate^\top \lxx \dstate + \lu^\top  \dcontrol + \frac{1}{2}\dcontrol^\top \luu \dcontrol + \mathbb{E}[V'(\state' + \dstate')]
\end{align*}

 Note that, when taking expectation for $\s^\top \dstate'$, the stochasticity and uncertainty terms disappear as the means of $\stochasticity$ and $\uncertainty$ are zero, while  for the second-order terms such as $\dstate'^\top (\cdot) \dstate'$, their covariances must be taken into account.\footnote{e.g., for a randomvariable $\xi\sim \mathcal{N}(0,\Sigma)$, $\mathbb{E}[\xi^\top S\xi ]= \tr (\Sigma S)$}. 

Expanding above equation with the definitions of partial derivatives,
\begin{align}\label{eq:bellman-stochastic}
\begin{split}
     \V(\state + \dstate)   &= \cost + s' +  \s'^\top (\A\dstate + \B\dcontrol) + \lx\dstate + \lu \dcontrol  \\
    &+  \frac{1}{2}\bigg((\A\dstate + \B\dcontrol)^\top \Ss'(A\dstate + B\dcontrol) +  \dstate^\top  \lxx \dstate + \dcontrol^\top  \luu \dcontrol \bigg) \\
    &+  \frac{1}{2} \big(\tr(\C\C^\top \Ss') + \tr(\D\D^\top \Ss')\big)\dt
    \end{split}
\end{align}
where the trace terms can be re-written as the following:
\begin{align*}
    \tr(\C\C^\top \Ss')&=\tr\bigg( \sum_i (\noisedyn^{(i)} + \noisedyn^{(i)}_\state + \noisedyn^{(i)}_\control)(\noisedyn^{(i)} + \noisedyn^{(i)}_\state + \noisedyn^{(i)}_\control)^\top \Ss'\bigg )\\
    &=  \sum_i (\noisedyn^{(i)} + \noisedyn^{(i)}_\state + \noisedyn^{(i)}_\control)^\top \Ss'(\noisedyn^{(i)} + \noisedyn^{(i)}_\state + \noisedyn^{(i)}_\control)
\end{align*}
where the superscripts $(i)$ denote the $i$th column of the corresponding matrix and subscripts denote their partial derivatives.

Then, combining the like terms together,  \eqref{eq:bellman-stochastic} becomes 
\begin{align*}
    &=\underbrace{l + s' +  \frac{1}{2}\dt \sum_i \noisedyn^{(i)\top } \Ss'\noisedyn^{(i)} + \frac{1}{2}\dt\sum_j \uncertaintydyn^{(j)\top} \Ss'\uncertaintydyn^{(j)}}_q\\
    &+ \underbrace{\bigg(\s'^\top \A + \lx^\top  + \dt\sum_i \noisedyn^{(i)\top }\Ss'\noisedyn^{(i)}_\state + \dt\sum_j \uncertaintydyn^{(j)\top }\Ss'\uncertaintydyn^{(j)}_\state \bigg)}_{\vi{q}^\top}\dstate\\
    &+\frac{1}{2} \dstate^\top \underbrace{\bigg(\A^\top \Ss'\A + \lxx +  \dt\sum_i \noisedyn^{(i)\top }_\state \Ss'\noisedyn^{(i)}_\state  + \dt\sum_j \uncertaintydyn^{(j)\top }_\state \Ss'\uncertaintydyn^{(j)}_\state  \bigg)}_{\vi{Q}}\dstate\\
    &+ \underbrace{\bigg( \s'^\top \B + \lu^\top  +  \dt\sum_i \noisedyn^{(i)\top }\Ss'\noisedyn^{(i)}_\control + \dt\sum_j \uncertaintydyn^{(j)\top }\Ss'\uncertaintydyn^{(j)}_\control\bigg)}_{\vi{g}^\top} \dcontrol    \\
    &+   \dstate^\top \underbrace{\bigg(\A^\top \Ss'\B +  \dt\sum_i \noisedyn^{(i)\top }_\state \Ss'\noisedyn^{(i)}_\control   + \dt\sum_j \uncertaintydyn^{(j)\top }_\state \Ss'\uncertaintydyn^{(j)}_\control   \bigg)}_{\vi{G}^\top}\dcontrol \\
    &+ \frac{1}{2}\dcontrol^\top  \underbrace{(\B^\top \Ss'\B + \luu +\dt\sum_i \noisedyn^{(i)\top }_\control \Ss'\noisedyn^{(i)}_\control   + \dt\sum_j \uncertaintydyn^{(j)\top }_\control \Ss'\uncertaintydyn^{(j)}_\control)}_{\vi{H}}\dcontrol  
\end{align*}

Minimizing this with respect to $\dcontrol$ gives the optimal $\dcontrol^*$:
\begin{align*}
    \dcontrol^* = -\vi{H}^{-1}\vi{g} -\vi{H}^{-1}\vi{G}\dstate,
\end{align*}

Plugging this $\dcontrol^*$ back into \eqref{eq:bellman-stochastic}, we get a quadratic form of $\V(\state + \dstate)$:
\begin{align*}
\V(\state + \dstate) = s + \dstate ^\top \s + \frac{1}{2}\dstate^\top \Ss\dstate 
\end{align*}
with 
\begin{align*}
    \Ss &= \vi{Q} - \G^\top \vi{H}^{-1}\G \\
    \s &= \q - \G^\top \vi{H}^{-\top} \vi{g} \\
    s &= q - \frac{1}{2} \g^\top \vi{H}^{-\top } \vi{g}
\end{align*}
Note that, in the absence of uncertainty terms, this is equivalent to iLQG as introduced in \cite{todorov2005generalized}, and further, in the absence of both uncertainty and stochsaticity, this is equivelent to iLQR~\cite{li2004iterative}.

To make a local improvement of the nominal trajectory, we perform a backward pass to update the value function and the optimal $\dcontrol^*$ for each timestep, starting with $\Ss_T = \lxx^{(T)},\,\s_T = \lx^{(T)},s_T = \cost_T$.

During a forward pass, the nominal trajectory is updated by applying  $\dcontrol = -\alpha  \vi{H}^{-1}\vi{g} -\vi{H}^{-1}\vi{G}\dstate$ to the deterministic part of the system. We use backtracking line-search to find $\alpha$ that minimizes the total cost.

For a long horizon task, Model Predictive Contol can be used with Robust-ILQG, which is to run Robust-ILQG for a short horizon repeatedly. We use this approach in our second experiment in \sref{subsec:Experiment-Quadrotor}.

\section{Experiments and Analysis}
We consider two simulated tasks: cart-pole swing-up and quadrotor control. For each task, one set of model parameters was used as the ``simulator'', and another set of model parameters was used as the ``real-world.'' Multiplicative noise was added to dynamics and Gaussian noise was added to observation. We compare GP-ILQG's performance with three optimal control algorithms: (1) ILQG using the ``real-world'' model, (2) ILQG using the incorrect ``simulator'' model, (3) Probabilistic DDP (PDDP)~\cite{pan2014probabilistic}, which is a variant of ILQG that relies only on data. For both GP-ILQG and PDDP, the same set of data and same GP implementation were used at each iteration.

As data gets large, computing Gaussian Process becomes computationally expensive, and thus it is common to use a subset of the data for computing the mean and covariance~\cite{rasmussen2006gaussian}. In our experiments, we keep all observations and uniformly sub-sample 300 data points at each iteration\footnote{For better results, more advanced techniques such as Sparse Pseudo-input GP~\cite{snelson2006sparse} or Sparse Spectral Gaussian Process~\cite{quia2010sparse} can be used.}. GP hyperparameters are optimized with the Gaussian Process for Machine Learning Toolbox~\cite{rasmussen2010gaussian}. If the learner's prediction error for a validation set is higher than that of its previous learner, we re-sample and re-train. 


\begin{figure*}[ht!]
\centering
\includegraphics{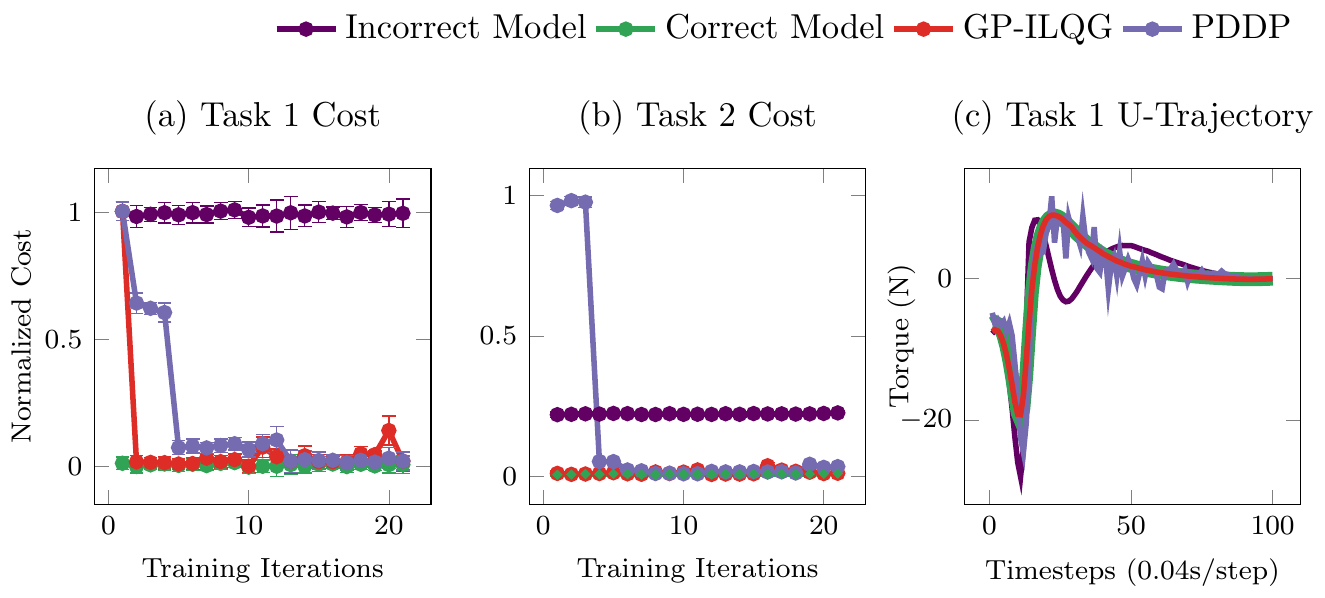}
    \caption{Cartpole swing-up.}
    \label{fig:cartpole}
\end{figure*}

\begin{figure*}[t!]
\centering
\includegraphics{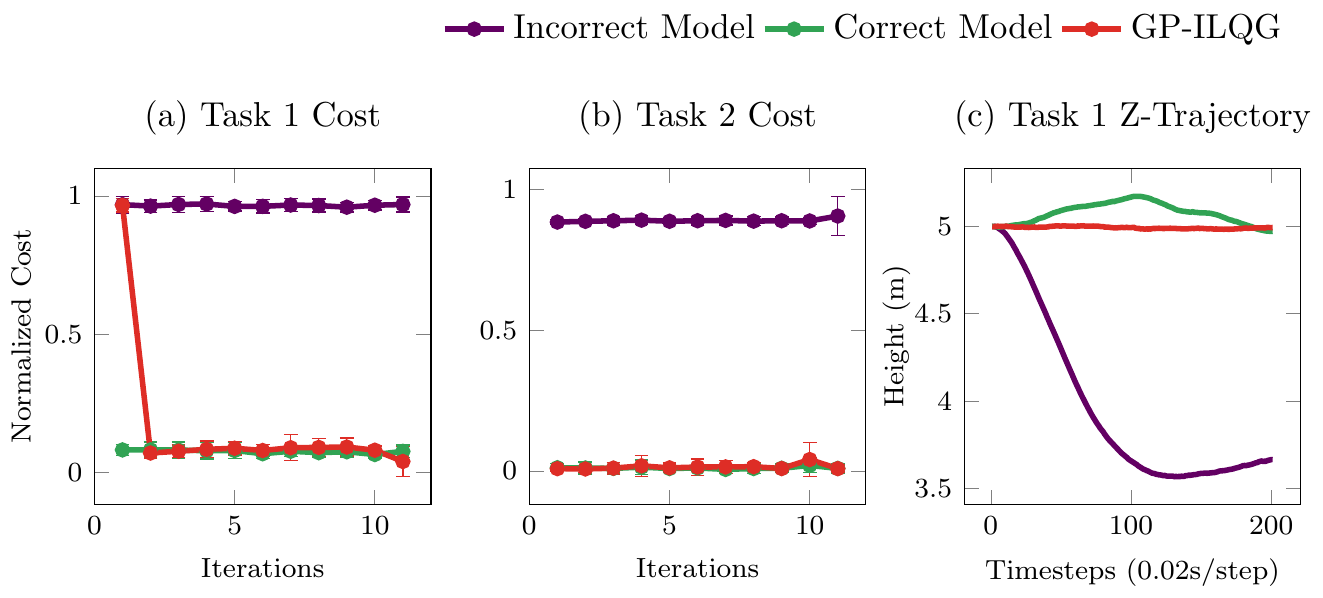}
\caption{Quadrotor control}
\label{fig:quadrotor}
\end{figure*}

\subsection{Cart-Pole Swing-Up}
In the cart-pole problem, the state is defined as $[x, \dot{x}, \theta, \dot{\theta}]$, where $x$ is the position of the cart along the $x-$axis and $\theta$ is the angle of the pole from the vertical downright position. Control input is the x-directional force (N).

The model parameters for the ``simulator'' and ``real-world'' model are given in \sref{sec:Appendix}. The real-world model has a 30\% longer pole.

We run two tasks in this experiment. In the first task, the initial state is $[0, \pi/4, 0, 0]$. \figref{fig:cartpole}(a) is the normalized cost for the first task. While both GP-ILQG and PDDP converges to the optimal performance, GP-ILQG is converges much quickly, within the first 2 iterations. 

The difference between GP-ILQG and PDDP is more noticeable in the second task (\figref{fig:cartpole}(b)), which starts from a different initial state $[0, -\pi/4, 0, 0]$. Both GP-ILQG and PDDP use the learner used in the previous task, but the initial cost for PDDP is significantly higher than GP-ILQG. We believe that this is because both algorithms explore an unexplored region in the first few iterations. While GP-ILQG relies on the simulator's inaccurate model in the unexplored region, PDDP has no information to make meaningful advancement until enough data is collected.

What is more noticeable is the improved performance of GP-ILQG over the simulator-based ILQG. The latter's suboptimal policy results in significantly higher cost in general. \figref{fig:cartpole}(c) shows the control sequences generated by the final policies of the four algorithms. GP-ILQG's control sequence is almost identical to the optimal sequence, and PDDP closely follows the optimal sequence as well. However, the simulator-based ILQG's control sequence is quite different due to its incorrect model.

\subsection{Quadrotor}\label{subsec:Experiment-Quadrotor}

We use the quadrotor model introduced in \cite{vanextended}. The model has 12-dimensional state, $\vi{x} = [\vi{p}, \vi{v}, \vi{r}, \vi{w}]^\top$, where $\vi{p}$~(m) and $\vi{v}$~(m/s) refers to the quadrotor's position and velocity in 3D space, $\vi{r}$ is orientation (rotation about axis $\vi{r}$ by angle $\|r\|$), and $\vi{w}$~(rad/s) is  angular velocity. It has 4 control inputs, $\vi{u} = [u_1, u_2, u_3,u_4]^\top$, which represent the force (N) exerted by the four rotors. The dynamics is given as the following:
\begin{align*}
    \dot{\vi{p}}  &= \vi{v}\\
    \dot{\vi{v}} &= -g \vi{e_3} + (\sum u_i) \exp([\vi{r}]\vi{e}_3 - k_v \vi{v})/m\\
    \dot{\vi{r}} &= \vi{w} + \frac{1}{2}[\vi{r}]\vi{w} + (1-\frac{1}{2} \|\vi{r}\|/\tan (\frac{1}{2}\|r\|)[\vi{r}]^2 /\|\vi{r}\|^2\\
    \dot{\vi{w}} &= J^{-1} (\rho (u_2 - u_4) \vi{e}_1 + \rho (u_3 - u_1) \vi{e}_2 \\
    & + k_m(u_1 - u_2 + u_3 - u_4) \vi{e}_3 - [\vi{w}]J \vi{w})
\end{align*}
where $\vi{e}_i$ are the standard basis vectors, $g = 9.8m/s^2$ is gravity, $k_v$ is the drag coefficient of rotors, $m$ (kg) is the mass, $J$ (kg $m^2$) is the moment of inertia matrix, and $\rho$ (m) is the distance between the center of mass and the center of rotors, and $k_m$ is a constant relating the force of rotors to its torque. $[\cdot]$ refers to the skew-symmetric cross product. The model parameters used are included in \sref{sec:Appendix}. The real-world model is 40\% heavier than the simulator's model.

We evaluate the performance of two tasks. The quadrotor starts at a position near $(0m, 0m, 5m)$ with zero velocity. In the first task, the goal is to move the quadrotor forward to $(4m, 0m, 5m)$ in 4 seconds. The second task is to drive the quadrotor to $(2m, 1m, 7m)$ in 4 seconds. The cost function was set to track a straight line from the initial position to the goal, with higher cost for maintaining the height.

In this experiment, we were not able to run PDDP to convergence with the same data used in GP-ILQG. We believe that this arises from the same problem we saw in the second task of cart-pole: PDDP has insufficient data to infer the unexplored state-space. We note that the original PDDP algorithm requires random trajectories as its initial data set instead of random variations of a single nominal trajectory. While our experiment does not indicate that PDDP is incapable of this task\footnote{A similar experiment with quadrotor control was shown to be successful in \cite{PanNIPSWS15}.}, it highlights our algorithm's data efficiency. Even with the small set of task-specific data, GP-ILQG converges in the first two iterations in the initial task (\figref{fig:quadrotor}(a) and converges immediately to the optimal policy in the second task (\figref{fig:quadrotor}(b)). \figref{fig:quadrotor}(c) compares the trajectories generated by the three algorithms. It shows that while our algorithm closely tracks the desired height, the simulator's suboptimal controller fails to recover from the vertical drop due to its incorrect mass model. 

\section{Conclusion}
In this paper, we proposed a novel algorithm that combines real-world data with a simulator's model to improve real-world performance of simulation-based optimal control. Our approach uses a Gaussian Process to correct a simulator's nonlinear model bias beyond the scope of its model space while incorporating the uncertainty of our estimate in computing a robust optimal control policy. Through simulated experiments, we have shown that our approach converges to the optimal performance within a few iterations and is capable of generalizing the learned dynamics for new tasks. 

Although our algorithm is capable of correcting significant model errors, it is limited by the quality of the initial policy based on the simulator's incorrect model. For example, the simulator's model can be sufficiently different from the true model such that the initial policy results in catastrophic damage to the robot. Our algorithm is incapable of measuring this initial uncertainty, although it can be improved by providing an initial set of expert-generated trajectories. We leave this as a future research direction.

\section{Appendix}\label{sec:Appendix}

\subsection{Stochasticity and uncertainty terms}
We define the partial derivatives for stochasticity and uncertainty terms as the following:

\begin{align}
    \noisedyn_\state\otimes \dstate &\triangleq \begin{bmatrix}
    \noisedyn^{(1)}_\state\dstate, ...,  \noisedyn^{(p)}_\state\dstate \end{bmatrix} \\
    \uncertaintydyn_\state \otimes \dstate &\triangleq \begin{bmatrix}
    \uncertaintydyn^{(1)}_\state\dstate, ...,  \uncertaintydyn^{(n)}_\state\dstate
    \end{bmatrix}
\end{align}
where $\noisedyn^{(i)}_\state, \uncertaintydyn^{(j)}_\state$ refer to the partial derivatives of $i, j$-th columns of $\noisedyn$ and $\uncertaintydyn$ with respect to $\state$.

\subsection{Models used in experiments}

We have used the following model parameters for the cart-pole and quadrotor experiments. ``Simulator'' refers to the incorrect model in the simulator. ``Real World'' is the model used during rollouts and policy evaluation; its model parameters are not revealed to our algorithm.

\subsubsection{Cartpole}
\begin{center}
  \begin{tabular}{rll}
  \toprule
     & Simulator & Real World \\ \midrule
    Cart Mass & 1 kg & 1 kg\\
    Pole Mass & 1 kg& 1 kg \\
    Pole Length &  {\bf 1 m}  & {\bf 1.3 m}\\ \bottomrule
  \end{tabular}
\end{center}

\subsubsection{Quadrotor}

$k_v$ is a constant relating the velocity to an opposite force, caused by rotor drag and induced inflow. $m$ (kg) is the mass, $J$ (kg $m^2$) is the moment of inertia matrix,  $\rho$ (m) is the distance between the center of mass and the center of the rotors.
\begin{center}
  \begin{tabular}{rll}
  \toprule
    & Simulator & Real World \\ \midrule
    $k_v$ & 0.15 & 0.15\\
    $k_m$ & 0.025 & 0.025\\
    m & {\bf 0.5} & {\bf 0.7} \\
    $J$ &  0.05 I  & 0.05 I \\
    $\rho$&0.17 & 0.17 \\ \bottomrule
  \end{tabular}
\end{center}
\bibliography{ref.bib}
\end{document}